\documentclass[10pt,twocolumn,letterpaper]{article}

\usepackage[pagenumbers]{cvpr} % To force page numbers, e.g. for an arXiv version

\usepackage[dvipsnames]{xcolor}
\usepackage{pifont}

\definecolor{darkred}{rgb}{0.7,0.1,0.1}
\definecolor{darkgreen}{rgb}{0.1,0.6,0.1}

\newlength\savewidth\newcommand\shline{\noalign{\global\savewidth\arrayrulewidth\global\arrayrulewidth1.25pt}\hline\noalign{\global\arrayrulewidth\savewidth}}
\newcommand{\tablestyle}[2]{\setlength{\tabcolsep}{#1}\renewcommand{\arraystretch}{#2}\centering\small}

\newcommand{\frameworkName}{DreamVideo\xspace}
\definecolor{cvprblue}{rgb}{0.21,0.49,0.74}
\usepackage[pagebackref,breaklinks,colorlinks,citecolor=cvprblue]{hyperref}
\newcommand{\tabincell}[2]{\begin{tabular}{@{}#1@{}}#2\end{tabular}}

\title{DreamVideo: Composing Your Dream Videos with \\Customized Subject and Motion}

\author{Yujie Wei$^{1\dagger}$\qquad Shiwei Zhang$^{2}$\qquad Zhiwu Qing$^{3\dagger}$\qquad Hangjie Yuan$^{4\dagger}$\qquad Zhiheng Liu$^{2\dagger}$ \\ Yu Liu$^{2}$\qquad Yingya Zhang$^{2}$\qquad Jingren Zhou$^{2}$\qquad Hongming Shan$^{1*}$
\\
$^{1}$ Fudan University\qquad
$^{2}$ Alibaba Group\\
$^{3}$ Huazhong University of Science and Technology\qquad
$^{4}$ Zhejiang University\\
{\tt\small yjwei22@m.fudan.edu.cn},\quad
{\tt\small qzw@hust.edu.cn},\quad {\tt\small  hj.yuan@zju.edu.cn},\quad {\tt\small hmshan@fudan.edu.cn},\\
{\tt\small \{zhangjin.zsw, pingzhi.lzh, ly103369, yingya.zyy, jingren.zhou\}@alibaba-inc.com}\\
}

\usepackage[misc]{ifsym}
\newcommand\blfootnote[1]{
    \begingroup
    \renewcommand\thefootnote{}\footnote{#1}
    \addtocounter{footnote}{-1}
    \endgroup
}
\usepackage[hang]{footmisc}

\begin{document}
\twocolumn[{
\renewcommand\twocolumn[1][]{#1}
\maketitle
\begin{center}
    \vspace{-10pt}
    \includegraphics[width=1.0\linewidth]{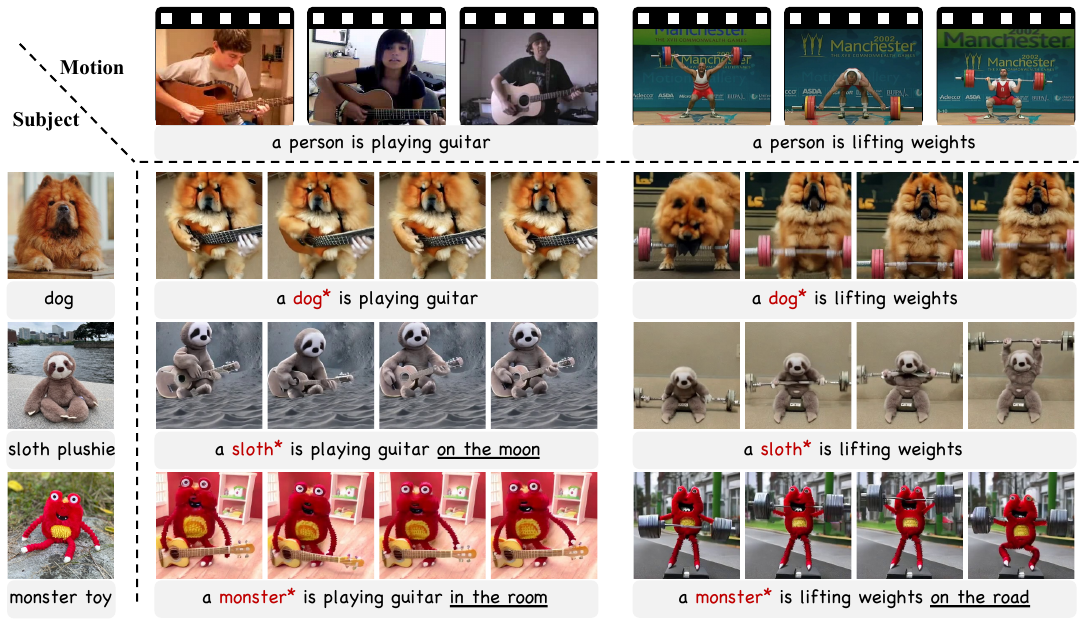}
    \vspace{-10pt}
    \captionsetup{type=figure}
    \caption{
        \textbf{Customized video generation results} of our proposed DreamVideo with specific subjects (left) and motions (top). Our method can customize \textit{both} subject identity and motion pattern to generate desired videos with various context descriptions.
    }
    \label{fig:teaser}
    % \vspace{6pt}
\end{center}
}]
{
    \blfootnote{
        $^*$Corresponding author.\\
        $^\dagger$Work done during the internships at Alibaba Group.\\
        This work is supported by Alibaba DAMO Academy through Alibaba Research Intern Program.
        }
}

\begin{abstract}
Customized generation using diffusion models has made impressive progress in image generation, but remains unsatisfactory in the challenging video generation task, as it requires the controllability of both subjects and motions.
To that end, we present \frameworkName, a novel approach to generating personalized videos from a few static images of the desired subject and a few videos of target motion.
\frameworkName decouples this task into two stages, subject learning and motion learning, by leveraging a pre-trained video diffusion model.
The subject learning aims to accurately capture the fine appearance of the subject from provided images, which is achieved by combining textual inversion and fine-tuning of our carefully designed identity adapter.
In motion learning, we architect a motion adapter and fine-tune it on the given videos to effectively model the target motion pattern.
Combining these two lightweight and efficient adapters allows for flexible customization of any subject with any motion.
Extensive experimental results demonstrate the superior performance of our \frameworkName over the state-of-the-art methods for customized video generation.
Our project page is at \url{https://dreamvideo-t2v.github.io}.
\end{abstract}   
\vspace{-10pt}
\section{Introduction}
\label{sec:intro}
The remarkable advances in diffusion models~\cite{DDPM, DDIM, stableDiffusion, DPM, DPMSolver} have empowered designers to generate photorealistic images and videos based on textual prompts, paving the way for customized content generation~\cite{textInversion, dreambooth}.
While customized image generation has witnessed impressive progress~\cite{customDiffusion, cones, cones2, break-a-scene}, the exploration of customized video generation remains relatively limited.
The main reason is that videos have diverse spatial content and intricate temporal dynamics simultaneously, presenting a highly challenging task to concurrently customize these two key factors.

Current existing methods~\cite{dreamix, tune-a-video} have effectively propelled progress in this field, 
but they are still limited to optimizing a single aspect of videos, namely spatial subject or temporal motion.
For example, Dreamix~\cite{dreamix} and Tune-A-Video~\cite{tune-a-video} optimize the spatial parameters and spatial-temporal attention to inject a subject identity and a target motion, respectively.
However, focusing only on one aspect (\ie, subject or motion) may reduce the model's generalization on the other aspect.
On the other hand, AnimateDiff~\cite{animatediff} trains temporal modules appended to the personalized text-to-image models for image animation.
It tends to pursue generalized video generation but suffers from a lack of motion diversity, such as focusing more on camera movements, making it unable to meet the requirements of customized video generation tasks well.
Therefore, we believe that effectively modeling both spatial subject and temporal motion is necessary to enhance video customization.

The above observations drive us to propose the \frameworkName, which can synthesize videos featuring the user-specified subject endowed with the desired motion from a few images and videos respectively, as shown in Fig.~\ref{fig:teaser}.
\frameworkName decouples video customization into subject learning and motion learning, which can reduce model optimization complexity and increase customization flexibility.
In subject learning, we initially optimize a textual identity using Textual Inversion~\cite{textInversion} to represent the coarse concept, and then train a carefully designed identity adapter with the frozen textual identity to capture fine appearance details from the provided static images.
In motion learning, we design a motion adapter and train it on the given videos to capture the inherent motion pattern.
To avoid the shortcut of learning appearance features at this stage, we incorporate the image feature into the motion adapter to enable it to concentrate exclusively on motion learning.
Benefiting from these two-stage learning, \frameworkName can flexibly compose customized videos with any subject and any motion once the two lightweight adapters have been trained.

To validate \frameworkName, we collect 20 customized subjects and 30 motion patterns as a substantial experimental set.
The extensive experimental results unequivocally showcase its remarkable customization capabilities surpassing the state-of-the-art methods.

In summary, our main contributions are:
\begin{enumerate}
    \item We propose \frameworkName, a novel approach for customized video generation with any subject and motion.
    To the best of our knowledge, this work makes the first attempt to customize \emph{both} {subject} identity and {motion}.
    \item We propose to decouple the learning of subjects and motions by the devised identity and motion adapters, which can greatly improve the flexibility of customization.
    \item We conduct extensive qualitative and quantitative experiments, demonstrating the superiority of \frameworkName over the existing state-of-the-art methods.
\end{enumerate}

\begin{figure*}[htp]
  \centering
  \includegraphics[width=1.0\linewidth]{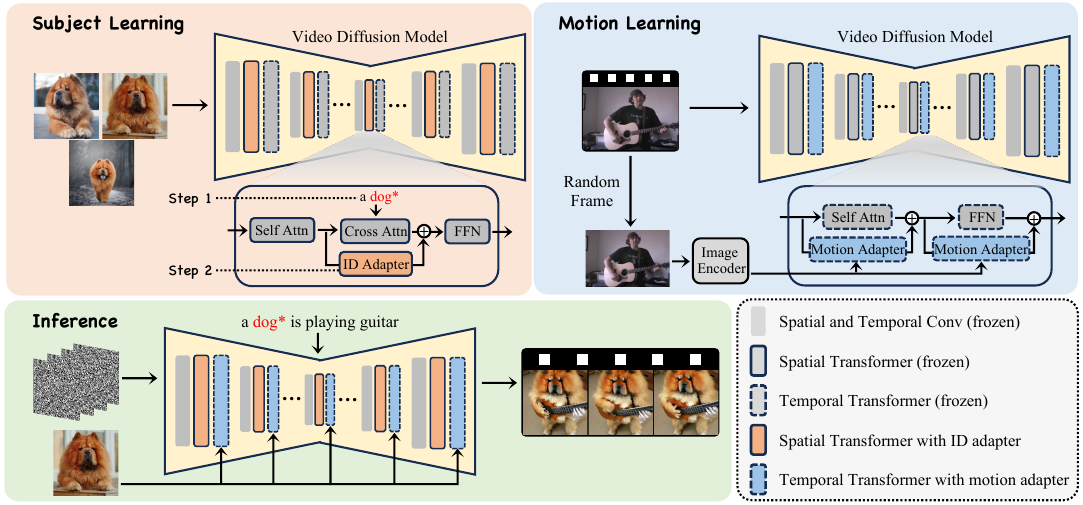}
  \caption{\textbf{Illustration of the proposed \frameworkName}, which decouples customized video generation into two stages. In subject learning, we first optimize a unique textual identity for the subject, and then train the devised identity adapter (ID adapter) with the frozen textual identity to capture fine appearance details. In motion learning, we pass a randomly selected frame from its training video through the CLIP image encoder, and use its embedding as the appearance condition to guide the training of the designed motion adapter.
  Note that we freeze the pre-trained video diffusion model throughout the training process. During inference, we combine the two lightweight adapters and randomly select an image provided during training as the appearance guidance to generate customized videos.
  }
  \label{fig:framework}
\end{figure*}  
\section{Related Work}
\label{sec:related_work}
\noindent\textbf{Text-to-video generation.}
Text-to-video generation aims to generate realistic videos based on text prompts and has received growing attention~\cite{t2v1, text2video_zero, videofusion, t2v2, pix2video, videocrafter1, diffsynth, large, videogen}. Early works are mainly based on Generative Adversarial Networks (GANs)~\cite{gan1, Mocogan, mostgan, gan2, gan3, stylegan_v} or autoregressive transformers~\cite{auto_regressive1, cogvideo, auto_regressive2, videogpt}. Recently, to generate high-quality and diverse videos, many works apply the diffusion model to video generation~\cite{video_p2p, videofactory, gen1, latent_shift, controlvideo, preserve, latent, free_bloom, lamp, fatezero, lavie, show1, i2vgen_xl}. 
Make-A-Video~\cite{make-a-video} leverages the prior of the image diffusion model to generate videos without paired text-video data. 
Video Diffusion Models~\cite{VDM} and ImagenVideo~\cite{imagenVideo} model the video distribution in pixel space by jointly training from image and video data.
To reduce the huge computational cost, VLDM~\cite{align-your-latents} and MagicVideo~\cite{magicVideo} apply the diffusion process in the latent space, following the paradigm of LDMs~\cite{stableDiffusion}. 
Towards controllable video generation, ModelScopeT2V~\cite{modelScope} and VideoComposer~\cite{videoComposer} incorporate spatiotemporal blocks with various conditions and show remarkable generation capabilities for high-fidelity videos.
These powerful video generation models pave the way for customized video generation.

\noindent\textbf{Customized generation.}
Compared with general generation tasks, customized generation may better accommodate user preferences. Most current works focus on subject customization with a few images~\cite{disenbooth, svdiff, SuTI, elite, instantbooth, continual_diffusion, hyperdreambooth}.
Textual Inversion~\cite{textInversion} represents a user-provided subject through a learnable text embedding without model fine-tuning. DreamBooth~\cite{dreambooth} tries to bind a rare word with a subject by fully fine-tuning an image diffusion model.
Moreover, some works study the more challenging multi-subject customization task~\cite{mix_of_show, cones, fastcomposer, subjectDiffusion, customDiffusion, cones2, anydoor}.
Despite the significant progress in customized image generation, customized video generation is still under exploration.
Dreamix~\cite{dreamix} attempts subject-driven video generation by following the paradigm of DreamBooth. However, fine-tuning the video diffusion model tends to overfitting and generate videos with small or missing motions.
A concurrent work~\cite{motionDirector} aims to customize the motion from training videos. Nevertheless, it fails to customize the subject, which may be limiting in practical applications. 
In contrast, this work proposes \frameworkName to 
effectively generate customized videos with \emph{both} specific subject and motion. 

\noindent\textbf{Parameter-efficient fine-tuning.}
Drawing inspiration from the success of parameter-efficient fine-tuning (PEFT) in NLP~\cite{prefix-tuning, lora} and vision tasks~\cite{adaptformer, aim, visualPrompts, ip_adapter}, some works adopt PEFT for video generation and editing tasks due to its efficiency~\cite{simDA, st_adapter}. In this work, we explore the potential of lightweight adapters, revealing their superior suitability for customized video generation.  
\section{Methodology}
\label{sec:method}
In this section, we first introduce the preliminaries of Video Diffusion Models. 
We then present \frameworkName to showcase how it can compose videos with the customized subject and motion. 
Finally, we analyze the efficient parameters for subject and motion learning while describing training and inference processes for our \frameworkName.

\subsection{Preliminary: Video Diffusion Models}
\label{sec:background}
Video diffusion models (VDMs)~\cite{VDM, align-your-latents, modelScope, videoComposer} are designed for video generation tasks by extending the image diffusion models~\cite{DDPM, stableDiffusion} to adapt to the video data. VDMs learn a video data distribution by the gradual denoising of a variable sampled from a Gaussian distribution. This process simulates the reverse process of a fixed-length Markov Chain. Specifically, the diffusion model $\epsilon_{\theta}$ aims to predict the added noise $\epsilon$ at each timestep $t$ based on text condition $c$, where $t \in \mathcal{U}(0, 1)$. The training objective can be simplified as a reconstruction loss:
\begin{align}
    \mathcal{L} = \mathbb{E}_{z, c, \epsilon \sim \mathcal{N}(0, \mathbf{\mathrm{I}}), t}\left[\left\| \epsilon - \epsilon_\theta\left(z_t, \tau_\theta(c), t\right)\right\|_2^2\right],
\label{eq:diffusion_loss}
\end{align}
where $z \in \mathbb{R}^{B \times F \times H \times W \times C}$ is the latent code of video data with $B, F, H, W, C$ being batch size, frame, height, width, and channel, respectively. $\tau_\theta$ presents a pre-trained text encoder. A noise-corrupted latent code $z_t$ from the ground-truth $z_0$ is formulated as $z_t = \alpha_t z_0 + \sigma_t \epsilon$,
where $\sigma_t = \sqrt{1 - \alpha_t^2}$, $\alpha_t$ and $\sigma_t$ are hyperparameters to control the diffusion process. 
Following ModelScopeT2V~\cite{modelScope}, we instantiate $\epsilon_\theta(\cdot, \cdot, t)$ as a 3D UNet, where each layer includes a spatiotemporal convolution layer, a spatial transformer, and a temporal transformer, as shown in Fig.~\ref{fig:framework}.

\begin{figure}[tp]
  \centering
  \includegraphics[width=0.85\linewidth]{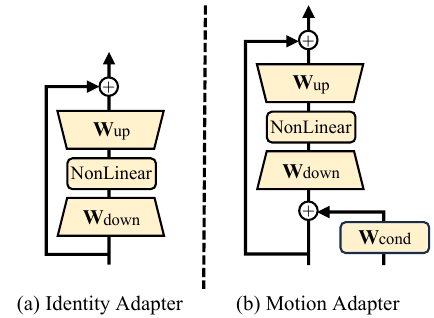}
  \caption{\textbf{Illustration of the devised adapters.} Both use a bottleneck structure. Compared to identity adapter, motion adapter adds a linear layer to incorporate the appearance guidance.}
  \label{fig:adapter}
\end{figure}

\subsection{\frameworkName}
Given a few images of one subject and multiple videos (or a single video) of one motion pattern, our goal is to generate customized videos featuring both the specific subject and motion. 
To this end, we propose \frameworkName, which decouples the challenging customized video generation task into subject learning and motion learning via two devised adapters, as illustrated in Fig.~\ref{fig:framework}. Users can simply combine these two adapters to generate desired videos.

\noindent\textbf{Subject learning.}
To accurately preserve subject identity and mitigate overfitting, we introduce a two-step training strategy inspired by~\cite{break-a-scene} for subject learning with 3$\sim$5 images, as illustrated in the upper left portion of Fig.~\ref{fig:framework}.

The first step is to learn a textual identity using Textual Inversion~\cite{textInversion}. We freeze the video diffusion model and only optimize the text embedding of pseudo-word ``$S^*$'' using Eq.~\eqref{eq:diffusion_loss}.
The textual identity represents the coarse concept and serves as a good initialization.

Leveraging only the textual identity is not enough to reconstruct the appearance details of the subject, so further optimization is required. 
Instead of fine-tuning the video diffusion model,
our second step is to learn a lightweight identity adapter by incorporating the learned textual identity. We freeze the text embedding and only optimize the parameters of the identity adapter. 
As demonstrated in Fig.~\ref{fig:adapter}(a), the identity adapter adopts a bottleneck architecture with a skip connection, which consists of a down-projected linear layer with weight $\mathbf{W}_{\mathrm {down}} \in \mathbb{R}^{l \times d}$, a nonlinear activation function $\sigma$, and an up-projected linear layer with weight $\mathbf{W}_{\mathrm {up}} \in \mathbb{R}^{d \times l}$, where $l > d$.
The adapter training process for the input spatial hidden state $h_t \in \mathbb{R}^{B \times (F \times h \times w) \times l}$ can be formulated as:
\begin{align}
    h_t^{\prime}=h_t + \sigma\left(h_t * \mathbf{W}_{\mathrm {down}}\right) * \mathbf{W}_{\mathrm{up}},
\end{align}
where $h, w, l$ are height, width, and channel of the hidden feature map, $h_t^{\prime}$ is the output of identity adapter, and $F=1$ because only image data is used. We employ GELU~\cite{gelu} as the activation function $\sigma$. 
In addition, we initialize $\mathbf{W}_{\mathrm{up}}$ with zeros to protect the pre-trained diffusion model from being damaged at the beginning of training~\cite{controlNet}.

\begin{figure}[tp]
  \centering
  \includegraphics[width=1.0\linewidth]{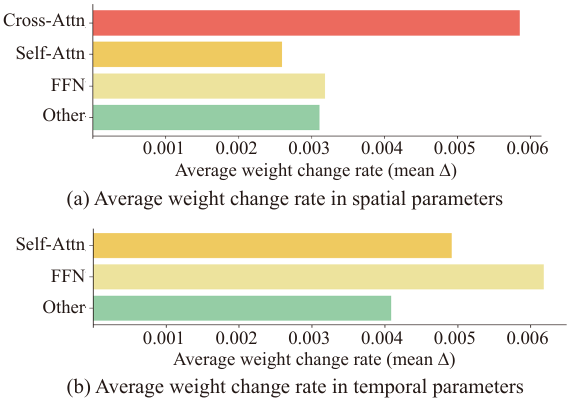}
  \caption{\textbf{Analysis of weight change} on updating all spatial or temporal model weights during fine-tuning. We observe that cross-attention layers play a key role in subject learning while the contributions of all layers are similar to motion learning.}
  \label{fig:average_change}
\end{figure}

\noindent\textbf{Motion learning.}
Another important property of customized video generation is to make the learned subject move according to the desired motion pattern from existing videos.
To efficiently model a motion, we devise a motion adapter with a structure similar to the identity adapter, as depicted in Fig.~\ref{fig:adapter}(b). 
Our motion adapter can be customized using a motion pattern derived from a class of videos (\eg, videos representing various dog motions), multiple videos exhibiting the same motion, or even a single video.

Although the motion adapter enables capturing the motion pattern, it inevitably learns the appearance of subjects from the input videos during training.
To disentangle spatial and temporal information, we incorporate appearance guidance into the motion adapter, forcing it to learn pure motion.
Specifically, we add a condition linear layer with weight $\mathbf{W}_{\mathrm {cond}} \in \mathbb{R}^{C^\prime \times l}$ to integrate appearance information into the temporal hidden state $\hat{h}_t \in \mathbb{R}^{(B \times h \times w) \times F \times l}$.
Then, we randomly select one frame from the training video and pass it through the CLIP~\cite{clip} image encoder to obtain its image embedding $e \in \mathbb{R}^{B \times 1 \times C^\prime}$. This image embedding is subsequently broadcasted across all frames, serving as the appearance guidance during training. The forward process of the motion adapter is formulated as:
\begin{align}
     &\hat{h}_t^e = \hat{h}_t + \mathrm{broadcast}(e * \mathbf{W}_{\mathrm {cond}}), \\
    &\hat{h}_t^{\prime} = \hat{h}_t + \sigma(\hat{h}_t^e * \mathbf{W}_{\mathrm {down}}) * \mathbf{W}_{\mathrm{up}},
\end{align}
where $\hat{h}_t^{\prime}$ is the output of motion adapter.
At inference time, we randomly take a training image provided by the user as the appearance condition input to the motion adapter.

\subsection{Model Analysis, Training and Inference}
\label{sec:train_inference}

\noindent\textbf{Where to put these two adapters.}
We address this question by analyzing the change of all parameters within the fine-tuned model to determine the appropriate position of the adapters. 
These parameters are divided into four categories: (1) cross-attention (only exists in spatial parameters), (2) self-attention, (3) feed-forward, and (4) other remaining parameters.
Following~\cite{weightChange, customDiffusion}, we use 
$\Delta_l=\left\|\theta_l^{\prime}-\theta_l\right\|_2 /\left\|\theta_l\right\|_2$ to calculate the weight change rate of each layer, where $\theta_l^{\prime}$ and $\theta_l$ are the updated and pre-trained model parameters of layer $l$. 
Specifically, to compute $\Delta$ of spatial parameters, we only fine-tune the spatial parameters of the UNet while freezing temporal parameters, for which the $\Delta$ of temporal parameters is computed in a similar way. 

We observe that the conclusions are different for spatial and temporal parameters. Fig.~\ref{fig:average_change}(a) shows the mean $\Delta$ of spatial parameters for the four categories when fine-tuning the model on ``Chow Chow'' images (dog in Fig.~\ref{fig:teaser}). The result suggests that the cross-attention layers play a crucial role in learning appearance compared to other parameters.
However, when learning motion dynamics in the ``bear walking'' video (see Fig.~\ref{fig:motion_compare}), all parameters contribute close to importance, as shown in Fig.~\ref{fig:average_change}(b). 
Remarkably, our findings remain consistent across various images and videos.
This phenomenon reveals the divergence of efficient parameters for learning subjects and motions.
Therefore, we insert the identity adapter to cross-attention layers while employing the motion adapter to all layers in temporal transformer. 

\noindent\textbf{Decoupled training strategy.}
Customizing the subject and motion simultaneously on images and videos requires training a separate model for each combination, which is time-consuming and impractical for applications.
Instead, we tend to decouple the training of subject and motion by optimizing the identity and motion adapters independently according to Eq.~\eqref{eq:diffusion_loss} with the frozen pre-trained model.

\noindent\textbf{Inference.}
During inference, we combine the two customized adapters and randomly select an image provided during training as the appearance guidance to generate customized videos. 
We find that choosing different images has a marginal impact on generated results.
Besides combinations, users can also customize the subject or motion individually using only the identity adapter or motion adapter.  
\section{Experiment}
\label{sec:exp}
\subsection{Experimental Setup}
\textbf{Datasets.}
For subject customization, we select subjects from image customization papers~\cite{dreambooth, cones2, mix_of_show} for a total of 20 customized subjects, including 9 pets and 11 objects.
For motion customization, we collect a dataset of 30 motion patterns from the Internet, the UCF101 dataset~\cite{ucf101}, the UCF Sports Action dataset~\cite{ucfSportsActions}, and the DAVIS dataset~\cite{davis}.
We also provide 42 text prompts used for extensive experimental validation, where the prompts are designed to generate new motions of subjects, new contexts of subjects and motions, and \etc.

\noindent\textbf{Implementation details.}
For subject learning, we take $\sim$3000 iterations for optimizing the textual identity following~\cite{textInversion, cones2}
with learning rate $1.0\times10^{-4}$, 
and $\sim$800 iterations for learning identity adapter
with learning rate $1.0\times10^{-5}$.
For motion learning, we train motion adapter for $\sim$1000 iterations with learning rate $1.0\times10^{-5}$. 
During inference, we employ DDIM~\cite{DDIM} with 50-step sampling and classifier-free guidance~\cite{classifier_free_guide} to 
generate 32-frame videos with 8 fps.
Additional details of our method and baselines are reported in Appendix~\ref{app:experimental_details}.

\begin{figure*}[htp]
  \centering
  \includegraphics[width=1.0\linewidth]{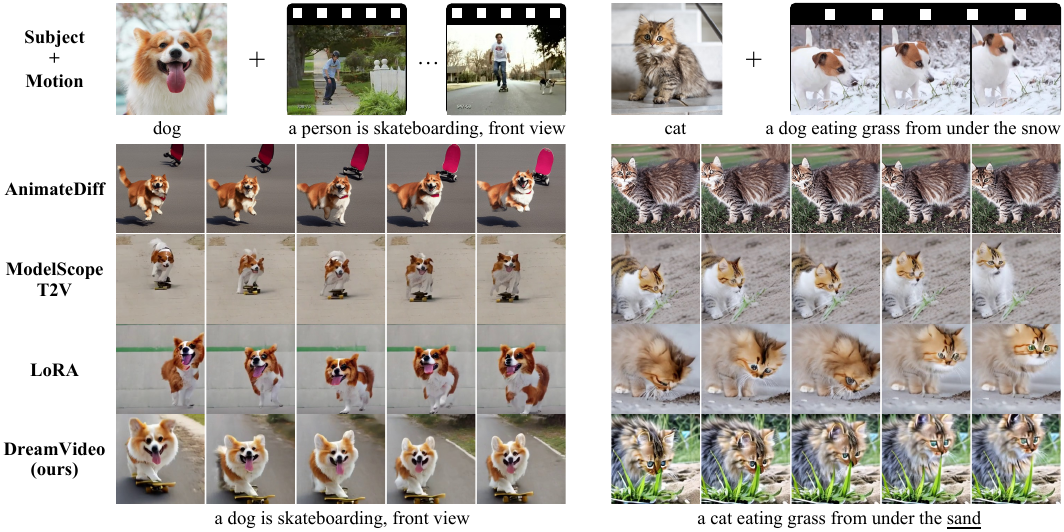}
  \caption{
  \textbf{Qualitative comparison of customized video generation with both subjects and motions}. \frameworkName accurately preserves both subject identity and motion pattern, while other methods suffer from fusion conflicts to some extent. Note that the results of AnimateDiff are generated by fine-tuning its provided pre-trained motion module and appending it to a DreamBooth~\cite{dreambooth} model.
  }
  \label{fig:flexibleCombination}
\end{figure*}
\begin{figure*}[htp]
  \centering
  \includegraphics[width=1.0\linewidth]{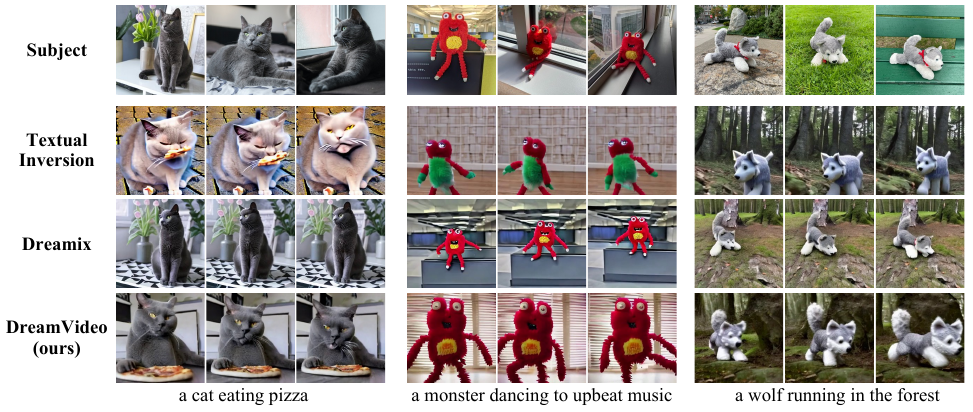}
  \caption{\textbf{Qualitative comparison of subject customization}. Our \frameworkName generates customized videos that preserve the precise subject appearance while conforming to text prompts with various contexts.
  }
  \label{fig:id_compare}
\end{figure*}
\begin{figure*}[htp]
  \centering
  \includegraphics[width=1.0\linewidth]{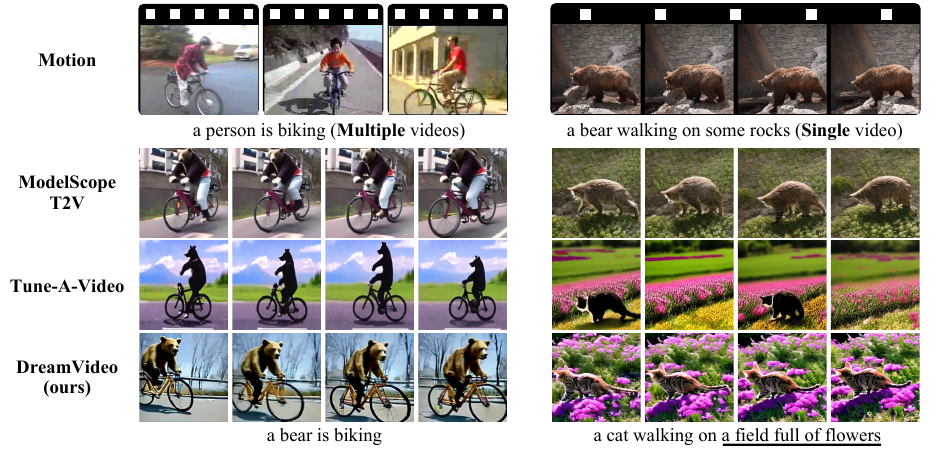}
  \caption{\textbf{Qualitative comparison of motion customization} between \frameworkName and other methods. Our approach effectively models specific motion patterns while avoiding appearance coupling, and generates temporal coherent as well as diverse videos.
  }
  \label{fig:motion_compare}
\end{figure*}

\noindent\textbf{Baselines.}
Since there is no existing work for customizing both subjects and motions, we consider comparing our method with three categories of combination methods: AnimateDiff~\cite{animatediff}, ModelScopeT2V~\cite{modelScope}, and LoRA fine-tuning~\cite{lora}. AnimateDiff trains a motion module appended to a pre-trained image diffusion model from Dreambooth~\cite{dreambooth}. 
However, we find that training from scratch leads to unstable results. For a fair comparison, we further fine-tune the pre-trained weights of the motion module provided by AnimateDiff and carefully adjust the hyperparameters.
For ModelScopeT2V and LoRA fine-tuning, we train spatial and temporal parameters/LoRAs of the pre-trained video diffusion model for subject and motion respectively, and then merge them during inference. 
In addition, we also evaluate our generation quality for customizing subjects and motions independently.
We evaluate our method against Textual Inversion~\cite{textInversion} and Dreamix~\cite{dreamix} for subject customization while comparing with Tune-A-Video~\cite{tune-a-video} and ModelScopeT2V for motion customization.

\noindent\textbf{Evaluation metrics.}
We evaluate our approach with the following four metrics, three for subject customization and one for video generation.
(1) \textit{CLIP-T} calculates the average cosine similarity between CLIP~\cite{clip} image embeddings of all generated frames and their text embedding.
(2) \textit{CLIP-I} measures the visual similarity between generated and target subjects. We compute the average cosine similarity between the CLIP image embeddings of all generated frames and the target images.
(3) \textit{DINO-I}~\cite{dreambooth}, another metric for measuring the visual similarity using ViTS/16 DINO~\cite{dino}. Compared to CLIP, the self-supervised training model encourages distinguishing features of individual subjects.
(4) \textit{Temporal Consistency}~\cite{gen1}, we compute CLIP image embeddings on all generated frames and report the average cosine similarity between all pairs of consecutive frames.

\begin{table}[h]
    \centering
    \resizebox{\columnwidth}{!}{
    \begin{tabular}{cccccc} 
        \textbf{Method} & \textbf{CLIP-T}& \textbf{CLIP-I}& \textbf{DINO-I} & \tabincell{c}{\textbf{T. Cons.} } & \textbf{Para.}\\ 
        \shline
         AnimateDiff~\cite{animatediff} & 0.298 & 0.657 & 0.432 & \textbf{0.982} & 1.21B\\
         ModelScopeT2V~\cite{modelScope} & 0.305 & 0.620 & 0.365 & 0.976 & 1.31B \\
         LoRA & 0.286 & 0.644 & 0.409 & 0.964 & 6M\\ 
         \hline
         \textbf{DreamVideo} & \textbf{0.314} & \textbf{0.665} & \textbf{0.452} & 0.971 & 85M\\ 
    \end{tabular}
    }
    \caption{\textbf{Quantitative comparison of customized video generation by combining different subjects and motions.} ``T. Cons.'' denotes Temporal Consistency. ``Para.'' means parameter number. }
    \label{tab:combine_compare}
\end{table}
\begin{table}[tp]
    \centering
    \resizebox{\columnwidth}{!}{
    \begin{tabular}{cccccc} 
        \textbf{Method} & \textbf{CLIP-T}& \textbf{CLIP-I}& \textbf{DINO-I} & \tabincell{c}{\textbf{T. Cons.}} & \textbf{Para.}\\ 
        \shline
         Textual Inversion~\cite{textInversion}& 0.278 & 0.668 & 0.362 & 0.961 & 1K \\
         Dreamix~\cite{dreamix}& 0.284 & \textbf{0.705} & 0.459 & \textbf{0.965} & 823M\\ 
         \hline
         \textbf{DreamVideo} & \textbf{0.295} & 0.701 & \textbf{0.475} & 0.964 & 11M\\ 
    \end{tabular}
    }
    \caption{\textbf{Quantitative comparison of subject customization.}
    }
    \label{tab:id_compare}
\end{table}

\subsection{Results}
In this section, we showcase results for both joint customization as well as individual customization of subjects and motions, further demonstrating the flexibility and effectiveness of our method.

\noindent\textbf{Arbitrary combinations of subjects and motions.}
We compare our \frameworkName with several baselines to evaluate the customization performance, as depicted in Fig.~\ref{fig:flexibleCombination}. We observe that AnimateDiff preserves the subject appearances but fails to model the motion patterns accurately, resulting in generated videos lacking motion diversity.
Furthermore, ModelScopeT2V and LoRA suffer from fusion conflicts during combination, where either subject identities are corrupted or motions are damaged.
In contrast, our \frameworkName achieves effective and harmonious combinations that the generated videos can retain subject identities and motions under various contexts; see Appendix~\ref{app:video_customization_more_results} for more qualitative results about combinations of subjects and motions.

Tab.~\ref{tab:combine_compare} shows quantitative comparison results of all methods. \frameworkName outperforms other methods across CLIP-T, CLIP-I, and DINO-I, which is consistent with the visual results.
Although AnimateDiff achieves highest Temporal Consistency, it tends to generate videos with small motions. In addition, our method remains comparable to Dreamix in Temporal Consistency but requires fewer parameters.

\begin{table}[t]
    \centering
    \tablestyle{8pt}{0.95}
    \begin{tabular}{cccc} 
    
        \textbf{Method} & \textbf{CLIP-T} & \textbf{T. Cons.} & \textbf{Para.}\\
         \shline
         ModelScopeT2V~\cite{modelScope}& 0.293 & 0.971 & 522M \\
         Tune-A-Video~\cite{tune-a-video}& 0.290 & 0.967 & 70M\\ 
         \hline
         \textbf{DreamVideo} & \textbf{0.309} & \textbf{0.975} & 74M\\
    \end{tabular}
    \caption{\textbf{Quantitative comparison of motion customization.} 
    }
    \label{tab:motion_compare}
\end{table}

\noindent\textbf{Subject customization.}
To verify the individual subject customization capability of our \frameworkName, we conduct qualitative comparisons with Textual Inversion~\cite{textInversion} and Dreamix~\cite{dreamix}, as shown in Fig.~\ref{fig:id_compare}. 
For a fair comparison, we employ the same baseline model, ModelScopeT2V, for all compared methods.
We observe that Textual Inversion makes it hard to reconstruct the accurate subject appearances.
While Dreamix captures the appearance details of subjects, the motions of generated videos are relatively small due to overfitting. Moreover, certain target objects in the text prompts, such as ``pizza'' in Fig.~\ref{fig:id_compare}, are not generated by Dreamix.
In contrast, our \frameworkName effectively mitigates overfitting and generates videos that conform to text descriptions while preserving precise subject appearances. 

The quantitative comparison for subject customization is shown in Tab.~\ref{tab:id_compare}.
Regarding the CLIP-I and Temporal Consistency, our method demonstrates a comparable performance to Dreamix while surpassing Textual Inversion.
Remarkably, our \frameworkName outperforms alternative methods in CLIP-T and DINO-I with relatively few parameters.
These results demonstrate that our method can efficiently model the subjects with various contexts.
Comparison with Custom Diffusion~\cite{customDiffusion} and more qualitative results are reported in Appendix~\ref{app:subject_customization_more_results}.

\begin{figure}[t]
  \centering
  \includegraphics[width=1.0\linewidth]{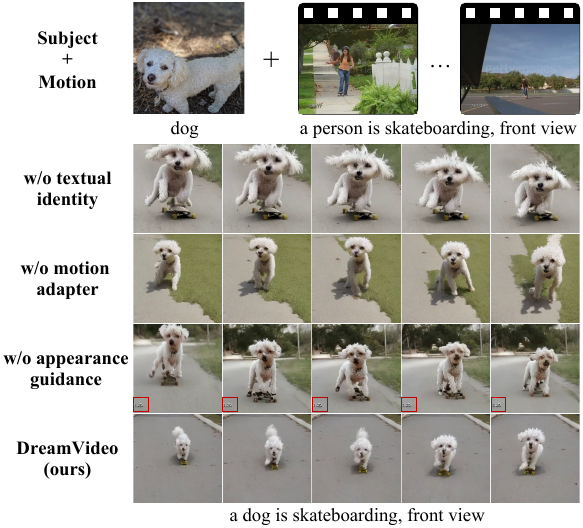}
  \caption{\textbf{Qualitative ablation studies} on each component.
  }
  \label{fig:ablation}
\end{figure}

\noindent\textbf{Motion customization.}
Besides subject customization, we also evaluate the motion customization ability of our \frameworkName by comparing it with several competitors, as shown in Fig.~\ref{fig:motion_compare}.
For a fair comparison, we only fine-tune the temporal parameters of ModelScopeT2V to learn a motion.
The results show that ModelScopeT2V inevitably fuses the appearance information of training videos, while Tune-A-Video suffers from discontinuity between video frames. 
In contrast, our method can capture the desired motion patterns while ignoring the appearance information of the training videos, generating temporally consistent and diverse videos; see Appendix~\ref{app:motion_customization_more_results} for more qualitative results about motion customization.

As shown in Tab.~\ref{tab:motion_compare}, our \frameworkName achieves the highest CLIP-T and Temporal Consistency compared to baselines, verifying the superiority of our method.

\noindent\textbf{User study.}
To further evaluate our approach, we conduct user studies for subject customization, motion customization, and their combinations respectively.
For combinations of specific subjects and motions, we ask 5 annotators to rate 50 groups of videos consisting of 5 motion patterns and 10 subjects. For each group, we provide 3$\sim$5 subject images and 1$\sim$3 motion videos; and compare our \frameworkName with three methods by generating videos with 6 text prompts.  We evaluate all methods with a majority vote from four aspects: Text Alignment, Subject Fidelity, Motion Fidelity, and Temporal Consistency.
Text Alignment evaluates whether the generated video conforms to the text description.
Subject Fidelity and Motion Fidelity measure whether the generated subject or motion is close to the reference images or videos. Temporal Consistency measures the consistency between video frames.
As shown in Tab.~\ref{tab:user_study}, our approach is most preferred by users regarding the above four aspects.
More details and user studies of subject customization as well as motion customization can be found in the Appendix~\ref{app:user_study}.

\begin{table}[t]
    \centering
    \resizebox{\columnwidth}{!}{
    \begin{tabular}{ccccc} 
    
        \textbf{Method} & \tabincell{c}{\textbf{Text}\\\textbf{Alignment}}& \tabincell{c}{\textbf{Subject}\\\textbf{Fidelity}}& \tabincell{c}{\textbf{Motion}\\\textbf{Fidelity}} & \tabincell{c}{\textbf{T.} \\\textbf{Cons.}}\\ 
        \shline
         ours \vs AD~\cite{animatediff} & 72.3 / 27.7 & 62.3 / 37.7 & 82.4 / 17.6 & 64.2 / 35.8 \\
         ours \vs MS~\cite{modelScope} & 62.4 / 37.6 & 66.2 / 33.8 & 56.6 / 43.4 & 51.5 / 48.5 \\
         ours \vs LoRA & 82.8 / 17.2 & 53.5 / 46.5 & 83.7 / 16.3 & 67.4 / 32.6 \\ 
    \end{tabular}
    }
    \caption{\textbf{Human evaluations} on customizing both subjects and motions between our method and alternatives. ``AD'' and ``MS'' are short for AnimateDiff and ModelScopeT2V, respectively.}
    \label{tab:user_study}
\end{table}

\begin{table}[t]
    \centering
    \resizebox{\columnwidth}{!}{
    \begin{tabular}{ccccc} 
        \textbf{Method} & \textbf{CLIP-T}& \textbf{CLIP-I}& \textbf{DINO-I} & \tabincell{c}{\textbf{T. Cons.}}\\ 
        \shline
         w/o textual identity & 0.310 & 0.657 & 0.435 & 0.968 \\
         w/o motion adapter & 0.295 & \textbf{0.701} & \textbf{0.475} & 0.964  \\
         w/o appearance guidance& 0.305 & 0.650 & 0.421 & 0.970 \\ 
         DreamVideo & \textbf{0.314} & 0.665 & 0.452 & \textbf{0.971} \\ 
    \end{tabular}
    }
    \caption{\textbf{Quantitative ablation studies} on each component.}
    \label{tab:ablation}
\end{table}

\subsection{Ablation Studies}
We conduct an ablation study on the effects of each component in the following. More ablation studies on the effects of parameter numbers and different adapters are reported in Appendix~\ref{app:ablation_study}.

\noindent\textbf{Effects of each component.} 
As shown in Fig.~\ref{fig:ablation}, we can observe that without learning the textual identity, the generated subject may lose some appearance details.
When only learning subject identity without our devised motion adapter, the generated video fails to exhibit the desired motion pattern due to limitations in the inherent capabilities of the pre-trained model.
In addition, without proposed appearance guidance, the subject identity and background in the generated video may be slightly corrupted due to the coupling of spatial and temporal information.
These results demonstrate each component makes contributions to the final performance. More qualitative results can be found in Appendix~\ref{app:ablation_more_qualitative}.

The quantitative results in Tab.~\ref{tab:ablation} show that all metrics decrease slightly without textual identity or appearance guidance, illustrating their effectiveness.
Furthermore, we observe that only customizing subjects leads to the improvement of CLIP-I and DINO-I, while adding the motion adapter can increase CLIP-T and Temporal Consistency. This suggests that the motion adapter helps to generate temporal coherent videos that conform to text descriptions.
\section{Conclusion}
\label{sec:conclusion}
In this paper, we present \frameworkName, a novel approach for customized video generation with any subject and motion. 
\frameworkName decouples video customization into subject learning and motion learning to enhance customization flexibility.
We combine textual inversion and identity adapter tuning to model a subject and train a motion adapter with appearance guidance to learn a motion.
With our collected dataset that contains 20 subjects and 30 motion patterns, we conduct extensive qualitative and quantitative experiments, demonstrating the efficiency and flexibility of our method in both joint customization and individual customization of subjects and motions. 

\noindent\textbf{Limitations.}
Although our method can efficiently combine a single subject and a single motion, it fails to generate customized videos that contain multiple subjects with multiple motions. One possible solution is to design a fusion module to integrate multiple subjects and motions, or to implement a general customized video model. We provide more analysis and discussion in Appendix~\ref{app:discussions}.

\clearpage
\renewcommand{\thetable}{A\arabic{table}}
\renewcommand{\thefigure}{A\arabic{figure}}
\renewcommand{\theequation}{A\arabic{equation}}
\setcounter{figure}{0}
\setcounter{table}{0}  
\setcounter{equation}{0}  

\appendix
\section*{Appendix}

\section{Experimental Details}
\label{app:experimental_details}
In this section, we supplement the experimental details of each baseline method and our method. 
To improve the quality and remove the watermarks of generated videos, we further fine-tune ModelScopeT2V~\cite{modelScope} for 30k iterations on a randomly selected subset from our internal data, which contains about 30,000 text-video pairs.
For a fair comparison, we use the fine-tuned ModelScopeT2V model as the base video diffusion model for all methods except for AnimateDiff~\cite{animatediff} and Tune-A-Video~\cite{tune-a-video}, both of which use the image diffusion model (Stable Diffusion~\cite{stableDiffusion}) in their official papers.
Here, we use Stable Diffusion v1-5\footnote{https://huggingface.co/runwayml/stable-diffusion-v1-5} as their base image diffusion model.
During training, unless otherwise specified, we default to using AdamW~\cite{adamW} optimizer with the default betas set to 0.9 and 0.999. The epsilon is set to the default $1.0\times10^{-8}$, and the weight decay is set to 0.
During inference, we use 50 steps of DDIM~\cite{DDIM} sampler and classifier-free guidance~\cite{classifier_free_guide} with a scale of 9.0 for all baselines. 
We generate 32-frame videos with 256 $\times$ 256 spatial
resolution and 8 fps.
All experiments are conducted using one NVIDIA A100 GPU.
In the following, we introduce the implementation details of baselines from subject customization, motion customization, and arbitrary combinations of subjects and motions (referred to as video customization).

\begin{table*}[htp]
    \centering
    \begin{tabular}{cccccc} 
        \textbf{Method} & \textbf{CLIP-T}& \textbf{CLIP-I}& \textbf{DINO-I} & \tabincell{c}{\textbf{T. Cons.}} & \textbf{Para.}\\ 
        \shline
         Custom Diffusion~\cite{customDiffusion}& 0.284 & 0.699 & 0.471 & 0.962 & 24M \\
         \textbf{DreamVideo (ours)} & \textbf{0.295} & \textbf{0.701} & \textbf{0.475} & \textbf{0.964} & 11M\\ 
    \end{tabular}
    \caption{\textbf{Quantitative comparison of subject customization between our method and Custom Diffusion~\cite{customDiffusion}.}
    ``T. Cons.'' and ``Para.'' denote Temporal Consistency and parameter number, respectively.
    }
    \label{tab:suppl_id_compare}
\end{table*}
\begin{table*}[ht]
    \centering
    \begin{tabular}{cccc} 
    
        \textbf{Method} & \tabincell{c}{\textbf{Text}\\\textbf{Alignment}}& \tabincell{c}{\textbf{Subject}\\\textbf{Fidelity}} & \tabincell{c}{\textbf{Temporal} \\\textbf{Consistency}}\\ 
        \shline
         ours \vs Textual Inversion~\cite{textInversion} & 58.4 / 41.6 & 79.8 / 20.2 & 69.7 / 30.3 \\
         ours \vs Dreamix~\cite{dreamix} &  63.7 / 36.3  & 56.0 / 44.0 & 50.8 / 49.2 \\
         ours \vs Custom Diffusion~\cite{customDiffusion} & 70.4 / 29.6 & 69.1 / 30.9 & 63.0 / 37.0 \\ 
    \end{tabular}
    \caption{\textbf{Human evaluations} on customizing subjects between our method and alternatives.}
    \label{tab:user_study_id}
\end{table*}
\begin{table*}[ht]
    \centering
    \begin{tabular}{cccc} 
    
        \textbf{Method} & \tabincell{c}{\textbf{Text}\\\textbf{Alignment}}& \tabincell{c}{\textbf{Motion}\\\textbf{Fidelity}} & \tabincell{c}{\textbf{Temporal} \\\textbf{Consistency}}\\ 
        \shline
         ours \vs ModelScopeT2V~\cite{modelScope} & 64.1 / 35.9 & 52.6 / 47.4 & 62.8 / 37.2 \\
         ours \vs Tune-A-Video~\cite{tune-a-video} & 73.8 / 26.2 & 52.4 / 47.6 & 74.1 / 25.9 \\
    \end{tabular}
    \caption{\textbf{Human evaluations} on customizing motions between our method and alternatives.}
    \label{tab:user_study_motion}
\end{table*}

\subsection{Subject Customization}
\label{app:subject_customization_details}
For all methods, we set batch size as 4 to learn a subject. 

\noindent\textbf{\frameworkName (ours).}
In subject learning, we take $\sim$3000 iterations for optimizing the textual identity following~\cite{textInversion, cones2}
with learning rate $1.0\times10^{-4}$, 
and $\sim$800 iterations for learning identity adapter
with learning rate $1.0\times10^{-5}$.
We set the hidden dimension of the identity adapter to be half the input dimension.
Our method takes $\sim$12 minutes to train the identity adapter on one A100 GPU.

\noindent\textbf{Textual Inversion~\cite{textInversion}.}
According to their official code\footnote{https://github.com/rinongal/textual\_inversion}, we reproduce Textual Inversion to the video diffusion model. We optimize the text embedding of pseudo-word ``$S^*$'' with prompt ``a $S^*$'' for 3000 iterations, and set the learning rate to $1.0\times10^{-4}$. 
We also initialize the learnable token with the corresponding class token.
These settings are the same as the first step of our subject-learning strategy.

\noindent\textbf{Dreamix~\cite{dreamix}.}
Since Dreamix is not open source, we reproduce its method based on the code\footnote{https://modelscope.cn/models/damo/text-to-video-synthesis} of ModelScopeT2V. According to the descriptions in the official paper, we only train the spatial parameters of the UNet while freezing the temporal parameters.
Moreover, we refer to the third-party implementation\footnote{\label{fn:dreamboothimpl}https://github.com/XavierXiao/Dreambooth-Stable-Diffusion} of DreamBooth~\cite{dreambooth} to bind a unique identifier with the specific subject. The text prompt used for target images is ``a [V] [category]'', where we initialize [V] with ``sks'', and [category] is a coarse class descriptor of the subject.
The learning rate is set to $1.0\times10^{-5}$, and the training iterations are 100 $\sim$ 200.

\noindent\textbf{Custom Diffusion~\cite{customDiffusion}.}
We refer to the official code\footnote{https://github.com/adobe-research/custom-diffusion} of Custom Diffusion and reproduce it on the video diffusion model. We train Custom Diffusion with the learning rate of $4.0\times10^{-5}$ and 250 iterations, as suggested in their paper. We also detach the start token embedding ahead of the class word with the text prompt ``a $S^*$ [category]''. We simultaneously optimize the parameters of the key as well as value matrices in cross-attention layers and text embedding of $S^*$. We initialize the token $S^*$ with the token-id 42170 according to the paper.

\subsection{Motion Customization}
\label{app:motion_customization_details}
To model a motion, we set batch size to 2 for training from multiple videos while batch size to 1 for training from a single video. 

\noindent\textbf{\frameworkName (ours).}
In motion learning, we train motion adapter for $\sim$1000 iterations with learning rate $1.0\times10^{-5}$. 
Similar to the identity adapter, the hidden dimension of the motion adapter is set to be half the input dimension.
On one A100 GPU, our method takes $\sim$15 and $\sim$30 minutes to learn a motion pattern from a single video and multiple videos, respectively. 

\noindent\textbf{ModelScopeT2V~\cite{modelScope}.}
We only fine-tune the temporal parameters of the UNet while freezing the spatial parameters. We set the learning rate to $1.0\times10^{-5}$, and also train 1000 iterations to learn a motion.

\noindent\textbf{Tune-A-Video~\cite{tune-a-video}.}
We use the official implementation\footnote{https://github.com/showlab/Tune-A-Video} of Tune-A-Video for experiments. 
The learning rate is $3.0\times10^{-5}$, and training iterations are 500.
Here, we adapt Tune-A-Video to train on both multiple videos and a single video.

\begin{table*}[ht]
    \centering
    \begin{tabular}{cccccc} 
        \textbf{Method} & \textbf{CLIP-T}& \textbf{CLIP-I}& \textbf{DINO-I} & \tabincell{c}{\textbf{T. Cons.} } & \textbf{Para.}\\ 
        \shline
         LoRA & 0.286 & 0.644 & 0.409 & 0.964 & 6M\\ 
         Adapter & \textbf{0.298} & \textbf{0.648} & \textbf{0.412} & \textbf{0.967} & 3M \\ 
    \end{tabular}
    \caption{\textbf{Quantitative comparison of video customization between Adapter and LoRA.} ``T. Cons.'' denotes Temporal Consistency. ``Para.'' means parameter number. }
    \label{tab:parameters_adapter_lora}
\end{table*}
\begin{table}[ht]
    \centering
    \begin{tabular}{ll|cc} 
    
        \textbf{Self-Attn} & \textbf{FFN} & \textbf{CLIP-T} & \textbf{T. Cons.}\\
         \shline
         Serial & Serial & 0.303 & 0.969 \\
         Serial & Parallel & 0.306 & 0.971 \\ 
         Parallel & Serial & 0.308 & \textbf{0.975} \\ 
         Parallel & Parallel & \textbf{0.309} & \textbf{0.975}\\
    \end{tabular}
    \caption{\textbf{Quantitative comparison of different adapters in motion customization.} ``Serial'' and ``Parallel'' mean using serial and parallel adapters in the corresponding layer, respectively.
    }
    \label{tab:different_adapters}
\end{table}

\subsection{Video Customization}

\noindent\textbf{\frameworkName (ours).}
We combine the trained identity adapter and motion adapter for video customization during inference. No additional training is required. We also randomly select an image provided during training as the appearance guidance. We find that choosing different images has a marginal impact on generated videos.

\noindent\textbf{AnimateDiff~\cite{animatediff}.}
We use the official implementation\footnote{https://github.com/guoyww/AnimateDiff} of AnimateDiff for experiments. 
AnimateDiff trains the motion module from scratch, but we find that this training strategy may cause the generated videos to be unstable and temporal inconsistent.
For a fair comparison, we further fine-tune the pre-trained weights of the motion module provided by AnimateDiff and carefully adjust the hyperparameters.
The learning rate is set to $1.0\times10^{-5}$, and training iterations are 50.
For the personalized image diffusion model, we use the third-party implementation\footref{fn:dreamboothimpl} code to train a DreamBooth model. During inference, we combine the DreamBooth model and motion module to generate videos.
 
\noindent\textbf{ModelScopeT2V~\cite{modelScope}.}
We train spatial/temporal parameters of the UNet while freezing other parameters to learn a subject/motion. 
Settings of training subject and motion are the same as Dreamix in Sec.~\ref{app:subject_customization_details} and ModelScopeT2V in Sec.~\ref{app:motion_customization_details}, respectively.
During inference, we combine spatial and temporal parameters into a UNet to generate videos.

\noindent\textbf{LoRA~\cite{lora}.}
In addition to fully fine-tuning, we also attempt the combinations of LoRAs. According to the conclusions in Sec.~3.3 of our main paper and the method of Custom Diffusion~\cite{customDiffusion}, we only add LoRA to the key and value matrices in cross-attention layers to learn a subject. For motion learning, we add LoRA to the key and value matrices in all attention layers. The LoRA rank is set to 32. Other settings are consistent with our \frameworkName. During inference, we merge spatial and temporal LoRAs into corresponding layers.

\section{More Results}
In this section, we conduct further experiments and showcase more results to illustrate the superiority of our \frameworkName.

\subsection{Video Customization}
\label{app:video_customization_more_results}
\begin{figure*}[htp]
  \centering
  \includegraphics[width=1.0\linewidth]{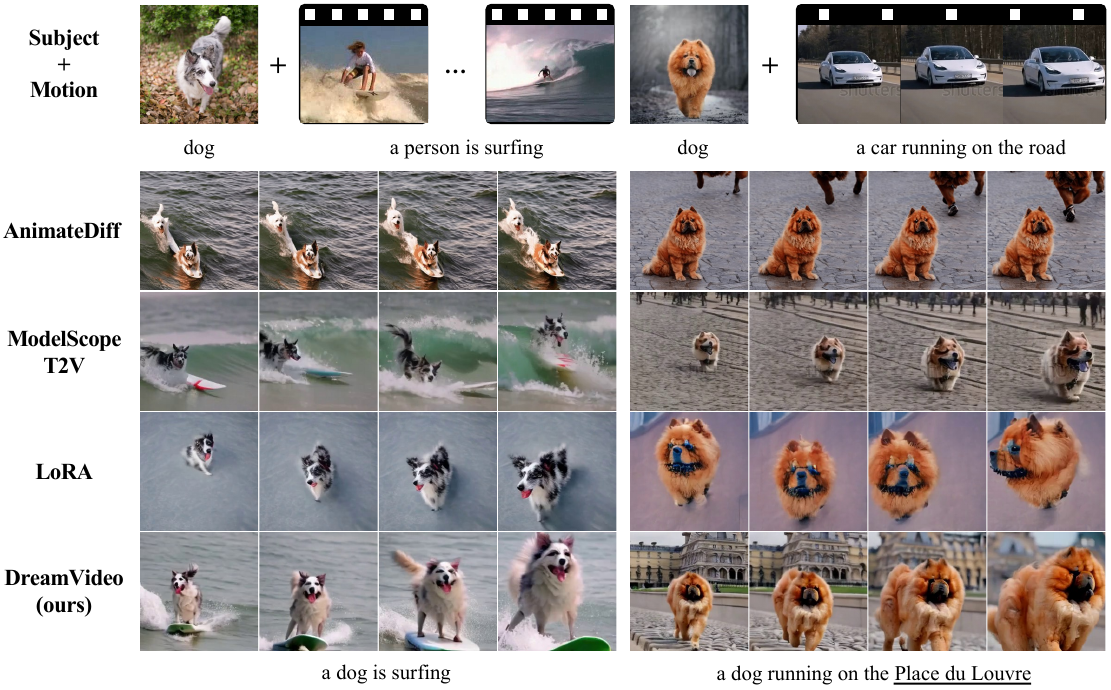}
  \caption{\textbf{Qualitative comparison of customized video generation with both subjects and motions.}
  }
  \label{fig:suppl_combine_compare}
\end{figure*}

We provide more results compared with the baselines, as shown in Fig.~\ref{fig:suppl_combine_compare}. The videos generated by AnimateDiff suffer from little motion, while other methods still struggle with the fusion conflict problem of subject identity and motion. In contrast, our method can generate videos that preserve both subject identity and motion pattern.

\subsection{Subject Customization}
\label{app:subject_customization_more_results}
\begin{figure*}[htp]
  \centering
  \includegraphics[width=1.0\linewidth]{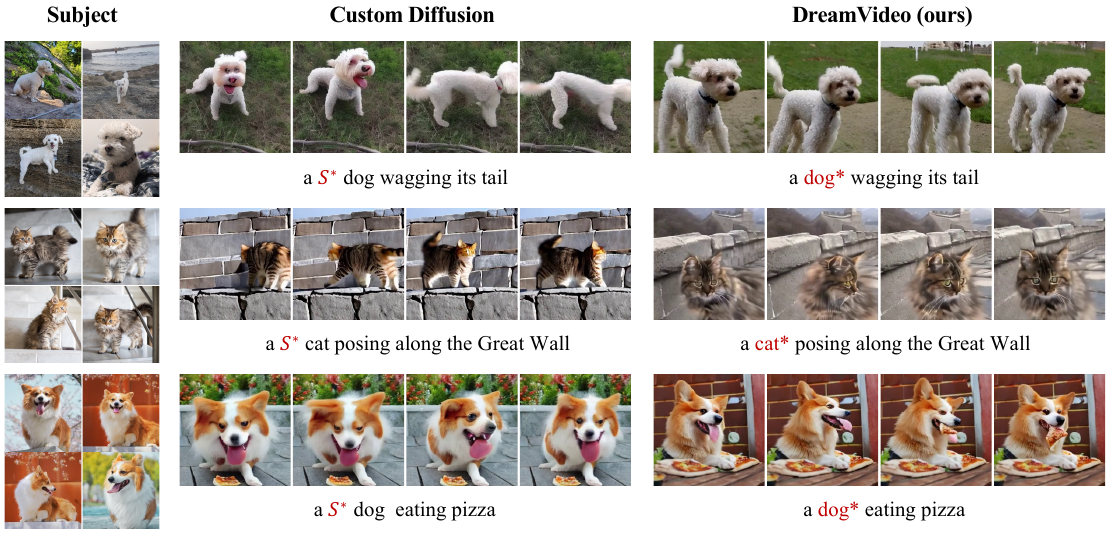}
  \caption{\textbf{Qualitative comparison of subject customization between our method and Custom Diffusion~\cite{customDiffusion}.}
  }
  \label{fig:suppl_id_compare}
\end{figure*}

In addition to the baselines in the main paper, we also compare our \frameworkName with another state-of-the-art method, Custom Diffusion~\cite{customDiffusion}.
Both the qualitative comparison in Fig.~\ref{fig:suppl_id_compare} and the quantitative comparison in Tab.~\ref{tab:suppl_id_compare} illustrate that our method outperforms Custom Diffusion and can generate videos that accurately retain subject identity and conform to diverse contextual descriptions with fewer parameters.

\begin{figure*}[htp]
  \centering
  \includegraphics[width=1.0\linewidth]{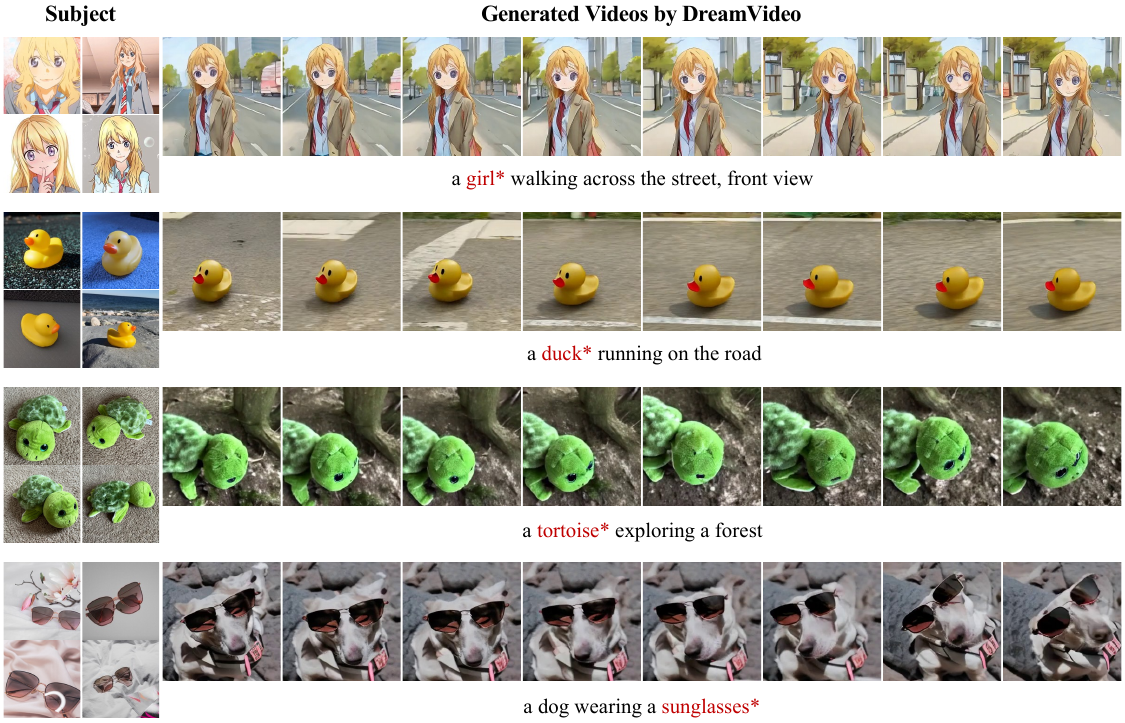}
  \caption{\textbf{More results of subject customization for Our \frameworkName}.
  }
  \label{fig:id_more_results}
\end{figure*}

As shown in Fig.~\ref{fig:id_more_results}, we provide the customization results for more subjects, further demonstrating the favorable generalization of our method.

\subsection{Motion Customization}
\label{app:motion_customization_more_results}
\begin{figure*}[htp]
  \centering
  \includegraphics[width=1.0\linewidth]{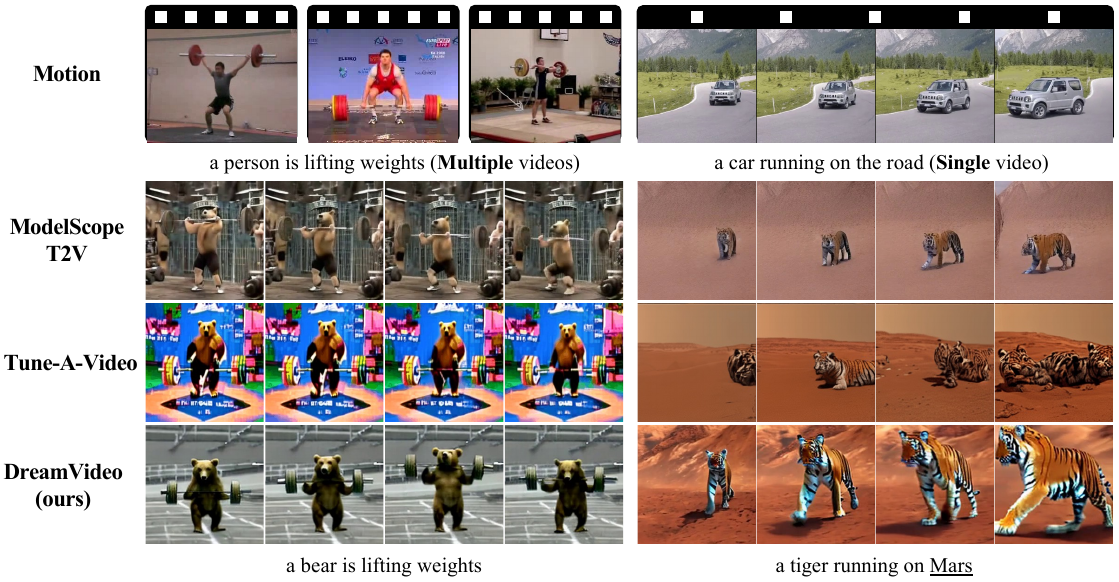}
  \caption{\textbf{Qualitative comparison of motion customization.}
  }
  \label{fig:suppl_motion_compare}
\end{figure*}

To further evaluate the motion customization capabilities of our method, we show more qualitative comparison results with baselines on multiple training videos and a single training video, as shown in Fig.~\ref{fig:suppl_motion_compare}. Our method exhibits superior performance than baselines and ignores the appearance information from training videos when modeling motion patterns.

\begin{figure*}[htp]
  \centering
  \includegraphics[width=1.0\linewidth]{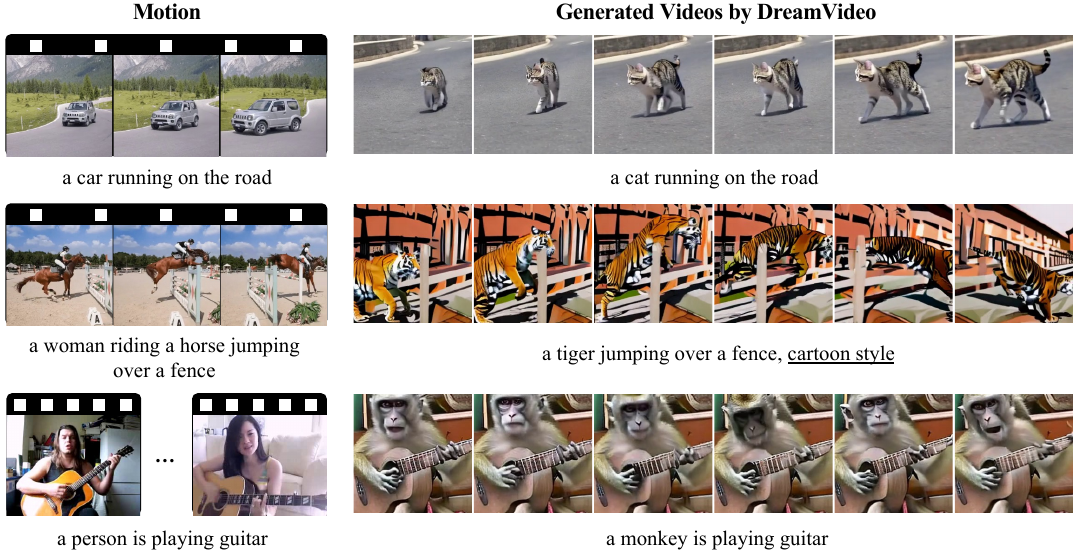}
  \caption{\textbf{More results of motion customization for Our \frameworkName}.
  }
  \label{fig:motion_more_results}
\end{figure*}

We showcase more results of motion customization in Fig.~\ref{fig:motion_more_results}, providing further evidence of the robustness of our method.

\subsection{User Study}
\label{app:user_study}
For subject customization, we generate 120 videos from 15 subjects, where each subject includes 8 text prompts. We present three sets of questions to participants with 3$\sim$5 reference images of each subject to evaluate Text Alignment, Subject Fidelity, and Temporal Consistency. 
Given the generated videos of two anonymous methods, we ask each participant the following questions: (1) Text Alignment: ``Which video better matches the text description?''; (2) Subject Fidelity: ``Which video's subject is more similar to the target subject?''; (3) Temporal Consistency: ``Which video is smoother and has less flicker?''.
For motion customization, we generate 120 videos from 20 motion patterns with 6 text prompts. We evaluate each pair of compared methods through Text Alignment, Motion Fidelity, and Temporal Consistency. The questions of Text Alignment and Temporal Consistency are similar to those in subject customization above, and the question of Motion Fidelity is like: ``Which video's motion is more similar to the motion of target videos?"
The human evaluation results are shown in Tab.~\ref{tab:user_study_id} and Tab.~\ref{tab:user_study_motion}. Our \frameworkName consistently outperforms other methods on all metrics.  

\section{More Ablation Studies}
\label{app:ablation_study}
\subsection{More Qualitative Results}
\label{app:ablation_more_qualitative}
\begin{figure*}[t]
  \centering
  \includegraphics[width=1.0\linewidth]{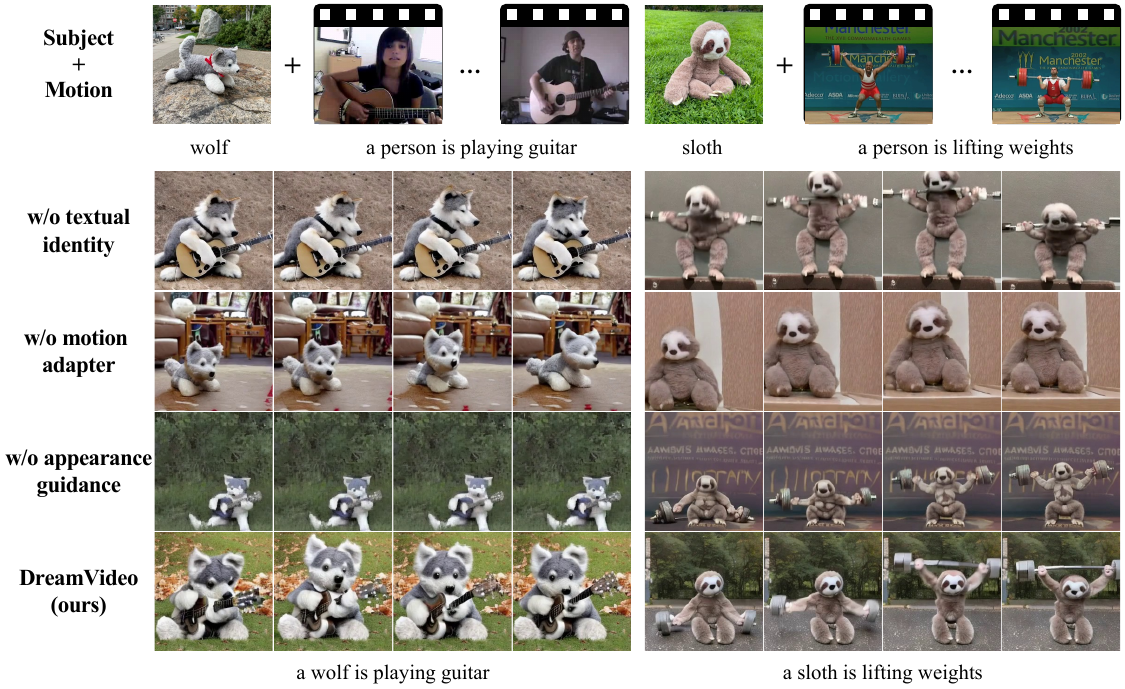}
  \caption{\textbf{Qualitative ablation studies} on each component.
  }
  \label{fig:suppl_ablation}
\end{figure*}

We provide more qualitative results in Fig.~\ref{fig:suppl_ablation} to further verify the effects of each component in our method.
The conclusions are consistent with the descriptions in the main paper.
Remarkably, we observe that without appearance guidance, the generated videos may learn some noise, artifacts, background, and other subject-unrelated information from training videos.

\subsection{Effects of Parameters in Adapter and LoRA}
To measure the impact of the number of parameters on performance, we reduce the hidden dimension of the adapter to make it have a comparable number of parameters as LoRA.
For a fair comparison, we set the hidden dimension of the adapter to 32 without using textual identity and appearance guidance.
We adopt the DreamBooth~\cite{dreambooth} paradigm for subject learning, which is the same as LoRA. Other settings are the same as our \frameworkName.

\begin{figure}[ht]
  \centering
  \includegraphics[width=1.0\linewidth]{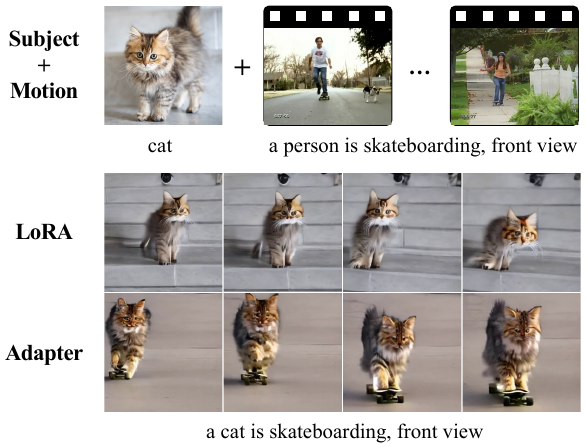}
  \caption{\textbf{Qualitative comparison of video customization between Adapter and LoRA}. The Adapter and LoRA here have the same hidden dimension (rank) and a comparable number of parameters.
  }
  \label{fig:suppl_adapter_lora_compare}
\end{figure}

As shown in Fig.~\ref{fig:suppl_adapter_lora_compare}, we observe that LoRA fails to generate videos that preserve both subject identity and motion. The reason may be that LoRA modifies the original parameters of the model during inference, causing conflicts and sacrificing performance when merging spatial and temporal LoRAs. In contrast, the adapter can alleviate fusion conflicts and achieve a more harmonious combination.

The quantitative comparison results in Tab.~\ref{tab:parameters_adapter_lora} also illustrate the superiority of the adapter compared to LoRA in video customization tasks.

\subsection{Effects of Different Adapters}
To evaluate which adapter is more suitable for customization tasks, we design 4 combinations of adapters and parameter layers for motion customization, as shown in Tab.~\ref{tab:different_adapters}. 
We consider the serial adapter and parallel adapter along with self-attention layers and feed-forward layers. 
The results demonstrate that using parallel adapters on all layers achieves the best performance.
Therefore, we uniformly employ parallel adapters in our approach.

\section{Social Impact and Discussions}
\label{app:discussions}
\noindent\textbf{Social impact.}
While training large-scale video diffusion models is extremely expensive and unaffordable for most individuals, video customization by fine-tuning only a few images or videos provides users with the possibility to use video diffusion models flexibly.
Our approach allows users to generate customized videos by arbitrarily combining subject and motion while also supporting individual subject customization or motion customization, all with a small computational cost. 
However, our method still suffers from the risks that many generative models face, such as fake data generation. 
Reliable video forgery detection techniques may be a solution to these problems.

\begin{figure}[ht]
  \centering
  \includegraphics[width=1.0\linewidth]{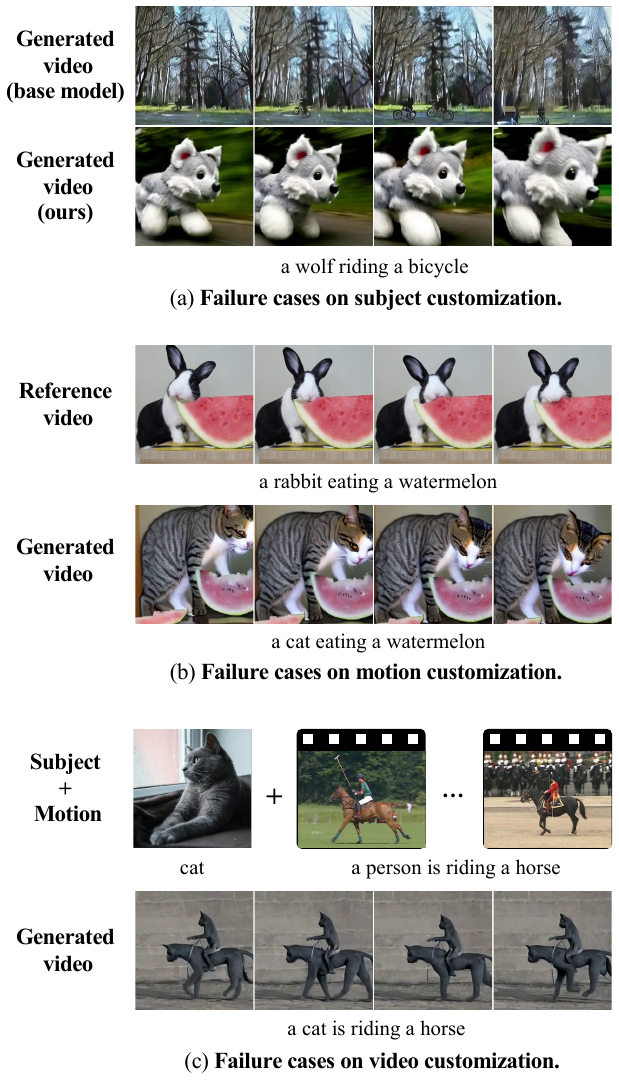}
  \caption{\textbf{Failure cases}. (a) Our method is limited by the inherent capabilities of the base model. (b) Our method may only learn the similar motion pattern on a fine single video motion. (c) Some difficult combinations that contain multiple objects still remain challenges.  
  }
  \label{fig:fail_cases}
\end{figure}

\noindent\textbf{Discussions.}
We provide some failure examples in Fig.~\ref{fig:fail_cases}. 
For subject customization, our approach is limited by the inherent capabilities of the base model. For example, in Fig.~\ref{fig:fail_cases}(a), the basic model fails to generate a video like ``a wolf riding a bicycle'', causing our method to inherit this limitation. The possible reason is that the correlation between ``wolf'' and ``bicycle'' in the training set during pre-training is too weak.
For motion customization, especially fine single video motion, our method may only learn the similar (rough) motion pattern and fails to achieve frame-by-frame correspondence, as shown in Fig.~\ref{fig:fail_cases}(b). Some video editing methods may be able to provide some solutions~\cite{tune-a-video, pix2video, stablevideo}.
For video customization, some difficult combinations that contain multiple objects, such as ``cat'' and ``horse'', still remain challenges. As shown in Fig.~\ref{fig:fail_cases}(c), our approach confuses ``cat'' and ``horse'' so that both exhibit ``cat'' characteristics. This phenomenon also exists in multi-subject image customization~\cite{customDiffusion}. One possible solution is to further decouple the attention map of each subject. 


\end{document}